%% file: acl_latex.tex
\documentclass[11pt]{article}

\usepackage[final]{acl}

\usepackage{times}
\usepackage{latexsym}

\usepackage[T1]{fontenc}
\usepackage[utf8]{inputenc}

\usepackage{microtype}

\usepackage{inconsolata}

\usepackage{graphicx}

\usepackage{booktabs}
\usepackage{multirow}
\usepackage{array}
\usepackage{amsmath}
\usepackage{amssymb}
\usepackage{algorithm}
\usepackage{algpseudocode}
\usepackage{colortbl}

\usepackage{tabularx}
\usepackage{ragged2e}
\usepackage{xcolor}
\definecolor{highlightorange}{RGB}{255,128,0}
\usepackage[most]{tcolorbox}
\tcbuselibrary{breakable}
\newcolumntype{Y}{>{\RaggedRight\arraybackslash}X}

\title{ESCRAG-R1: Retrieval-Augmented Reinforcement Learning for Emotional Support Conversation}

\author{
 \textbf{Weichu Liu\textsuperscript{1,2}$^{*}$},
 \textbf{Yuxuan Hu\textsuperscript{3}$^{*}$},
 \textbf{Yirong Sun\textsuperscript{4}},
 \textbf{Ningning Mao\textsuperscript{5}},
 \textbf{Ziyun Zhang\textsuperscript{1}},
 \\
 \textbf{Jian Chen\textsuperscript{6}},
 \textbf{Mingyang Xu\textsuperscript{2}},
 \textbf{Qishan Zhong\textsuperscript{2}},
 \textbf{Chengming Li\textsuperscript{2}$^{\dagger}$},
\\
 \textsuperscript{1} Beijing Institute of Technology,
 \textsuperscript{2} Shenzhen MSU-BIT University,
 \\
 \textsuperscript{3} City University of Hong Kong,
 \textsuperscript{4} Shenzhen University of Advanced Technology,
 \\
 \textsuperscript{5} Beijing Normal University,
 \textsuperscript{6} The University of Hong Kong
\\
 \small{
 liuwc@bit.edu.cn, yuxuanhu7-c@my.cityu.edu.hk, licm@smbu.edu.cn
 }
}

\begin{document}
\maketitle

\renewcommand{\thefootnote}{\fnsymbol{footnote}}
\setcounter{footnote}{0}
\footnotetext[1]{Equal contribution.}
\footnotetext[2]{Corresponding author.}

\input{chapter/0_abstract}
\input{chapter/1_Intro}
\input{chapter/2_Related_Work}

\input{chapter/3_Method}

\input{chapter/4_Experiment}
\input{chapter/5_Conclusion}
\input{chapter/Limitations_Statement}
\bibliography{custom}

\appendix
\input{chapter/appendix}

\end{document}

%% file: chapter/0_abstract.tex
\begin{abstract}
Emotional Support Conversation (ESC) systems aim to provide holistic support by balancing professional therapeutic competence with natural empathy. However, existing methods struggle to simultaneously achieve structured, stage-aware reasoning and seamless empathy-expertise alignment, often resulting in an artificial splicing of clinical strategies and generic reassurance. To overcome these limitations, we propose \textsc{\textbf{ESCRAG-R1}}, a unified framework that integrates retrieval-based psychological guidance into Group Relative Policy Optimization (GRPO). By incorporating retrieval into the reinforcement learning loop, ESCRAG-R1 transforms external knowledge into a robust learning signal that stimulates explicit internal reasoning prior to generation and fundamentally reshapes the model's internal policy. To provide the reliable supervision required for this optimization, we construct \textsc{\textbf{ESC-Preference}}, a high-quality dataset based on a Client--Counselor--Judge evaluation framework that delivers precise, empathy-aware reward signals. Extensive experiments demonstrate that ESCRAG-R1 significantly outperforms existing baselines by mitigating superficial splicing and realizing a natural integration of professional guidance and empathetic expression. Code and datasets are released at \url{https://github.com/Matcha-Liu/ESCRAG-R1}.
% Emotional Support Conversation (ESC) aims to provide responses that are both empathetic and psychologically grounded for users experiencing emotional distress. Existing ESC approaches generally fall into parameter-based and inference-augmented paradigms, but they often struggle to jointly achieve structured, stage-aware reasoning and empathy-expertise alignment. To address this gap, we propose \textsc{ESCRAG-R1}, a unified framework that incorporates retrieval-based strategy guidance into Group Relative Policy Optimization (GRPO) under professionally grounded and empathy-aware reward supervision, allowing psychologically relevant guidance to participate in both response generation and policy optimization. To support reward modeling and retrieval grounding, we construct \textsc{ESC-Preference}, a preference-aligned emotional support dataset based on a three-dimensional evaluation framework that jointly assesses professional competence and empathetic quality. Experimental results show that ESCRAG-R1 consistently improves the overall quality of emotional support responses. All of our code and datasets will be opensourced in the future.
\end{abstract}

%% file: chapter/1_Intro.tex
\section{Introduction}
\label{sec-intro}
% With the rapid advancement of large language models (LLMs) in contextual understanding, reasoning, and text generation \citep{yi2025survey}, an increasing number of users are turning to Emotional Support Conversation (ESC) systems for psychological comfort and guidance \citep{deepseekai2024deepseekv3technicalreport,openai2025gpt5,comanici2025gemini}. In real-world scenarios, users often share their experiences and emotional states with ESC systems, expecting empathetic and constructive responses \citep{rashkin2019towards,lin2019moel}. The development of ESC systems is progressively advancing toward deeper Emotional Intelligence, aiming to provide users with comprehensive emotional support that balances professional competence and empathy \citep{zhong2021care,wang2023emotional}.
\input{picture/banner}
Driven by growing psychological needs and LLM advancements \citep{yi2025survey,li2025ctr}, Emotional Support Conversation (ESC) systems have become vital resources for users seeking comfort and guidance \citep{deepseekai2024deepseekv3technicalreport,openai2025gpt5,comanici2025gemini}. In real-world scenarios, users often share their experiences and emotional states with ESC systems, expecting empathetic and constructive responses \citep{rashkin2019towards,lin2019moel}. The development of ESC systems is advancing toward deeper Emotional Intelligence, aiming to provide users with comprehensive emotional support that balances professional competence and empathy \citep{zhong2021care,wang2023emotional,chen2025mghft,hu2026emotion}.

% Building on this vision, effective ESC systems require two key capabilities. 
% The first is \textit{structured stage-aware reasoning}. Emotional support is inherently process-oriented: a high-quality response involves inferring the client’s counseling stage, assessing their psychological state, and selecting appropriate intervention strategies \citep{beck2020cognitive,hayes2006acceptance,elliott2002effectiveness}. Such reasoning reflects established therapeutic principles, where empathy and guidance emerge through coherent counseling processes rather than isolated responses. 
% The second is \textit{empathy-expertise alignment}. Psychologically grounded guidance should be expressed in a warm and non-lecturing manner. Responses that are emotionally warm but psychologically ungrounded may offer only superficial reassurance, while technically accurate yet affectively detached replies risk alienating vulnerable users \citep{nienhuis2018therapeutic}. Therefore, effective emotional support requires both psychologically grounded intervention reasoning and carefully calibrated empathetic language.

To achieve deep emotional intelligence, an effective ESC system must possess two core capabilities. The first is \textit{structured, stage-aware reasoning}. Because emotional support is inherently process-oriented, generating a high-quality response requires inferring the client’s counseling stage, assessing their psychological state, and selecting appropriate interventions \citep{beck2020cognitive,hayes2006acceptance,elliott2002effectiveness}. This explicit reasoning ensures that empathy and guidance emerge from a coherent framework rather than isolated, reactive utterances.
The second capability is \textit{empathy-expertise alignment}. Psychologically grounded guidance must be delivered in a supportive, non-didactic manner. Responses that are emotionally warm but clinically ungrounded offer only superficial reassurance, while technically accurate yet affectively detached replies risk alienating vulnerable users \citep{nienhuis2018therapeutic}. Ultimately, effective support hinges on seamlessly integrating rigorous psychological reasoning with carefully calibrated empathetic language.

% Existing research on ESC systems can generally be divided into two paradigms: parameter-based approaches and inference-augmented approaches. (i) Parameter-based approaches shape supportive behaviors by updating model parameters through supervised fine-tuning \citep{chen2023soulchat,zhang2024cpsycoun} or reinforcement learning \citep{zhao-etal-2025-chain,wang2025rlver} on counseling datasets. Although such training can internalize desirable response patterns, \textbf{it often lacks explicit guidance for structured, stage-aware reasoning}, making it difficult to provide appropriate emotional support across different counseling stages as the client’s psychological state evolves. (ii) Inference-augmented approaches improve response quality by incorporating counseling-oriented agent frameworks \citep{zhou2025diacbt,zhang2024escot} or retrieval-augmented generation \citep{hu2024aptness,guo2024soullmate}. While these methods introduce psychologically relevant knowledge during response generation, such knowledge typically \textbf{serves as auxiliary guidance at inference time rather than a learning signal that shapes the model’s internal policy}. As a result, the generated responses may mechanically combine counseling-related knowledge with generic reassurance, rather than naturally integrating professional support with empathetic expression.

Existing ESC research generally follows two paradigms, both struggling to simultaneously fulfill these ideal capabilities.
(i) Parameter-based approaches use supervised fine-tuning \citep{chen2023soulchat,zhang2024cpsycoun} or reinforcement learning \citep{zhao-etal-2025-chain,wang2025rlver} to mimic empathetic patterns. However, they critically lack explicit guidance for structured, stage-aware reasoning, relying on surface-level imitation rather than dynamically adapting to the client’s evolving psychological state.
(ii) Inference-augmented approaches incorporate agents \citep{zhou2025diacbt,zhang2024escot} or retrieval-augmented generation \citep{hu2024aptness,guo2024soullmate} to inject psychological knowledge. Yet, this knowledge typically serves merely as external auxiliary guidance at inference time, rather than a definitive learning signal reshaping the model’s internal policy.
Consequently, avoiding this artificial splicing requires a unified framework that integrates explicit reasoning guidance directly into the model's internal policy.

To overcome these limitations, we propose \textbf{\textsc{ESCRAG-R1}}, a unified framework that integrates retrieval-based psychological guidance into Group Relative Policy Optimization (GRPO), as shown in Figure~\ref{fig:banner}. \textsc{ESCRAG-R1} leverages retrieved counseling strategies to stimulate explicit, stage-aware reasoning prior to generation, ensuring the model conducts an internal psychological assessment rather than surface-level imitation. Furthermore, by incorporating this retrieval process into the GRPO loop, external knowledge acts as a robust learning signal that reshapes the model's internal policy, seamlessly integrating professional therapeutic strategies with empathetic expression. To provide reliable reward supervision for this optimization, we construct a Client--Counselor--Judge evaluation framework and build \textbf{\textsc{ESC-Preference}}. This dataset delivers precise reward signals for GRPO, while its preferred responses serve as practice-grounded exemplars in the retrieval corpus for both training and inference.

Our contributions are summarized as:
\begin{itemize}
    \item We propose \textsc{ESCRAG-R1}, a retrieval-augmented reinforcement learning framework that incorporates psychological guidance into GRPO for stage-aware emotional support.

\item We construct \textsc{ESC-Preference} based on a three-dimensional Client--Counselor--Judge evaluation framework, supporting both reward modeling and retrieval grounding.

\item Extensive experiments show that \textsc{ESCRAG-R1} improves empathy-expertise alignment and generates more grounded, supportive, and therapeutically appropriate responses.

\end{itemize}

%% file: picture/banner.tex
\begin{figure}[!t]
\setlength\abovecaptionskip{0.2\baselineskip}
\setlength\belowcaptionskip{0.2\baselineskip}
\centering
\includegraphics[width=0.45\textwidth]{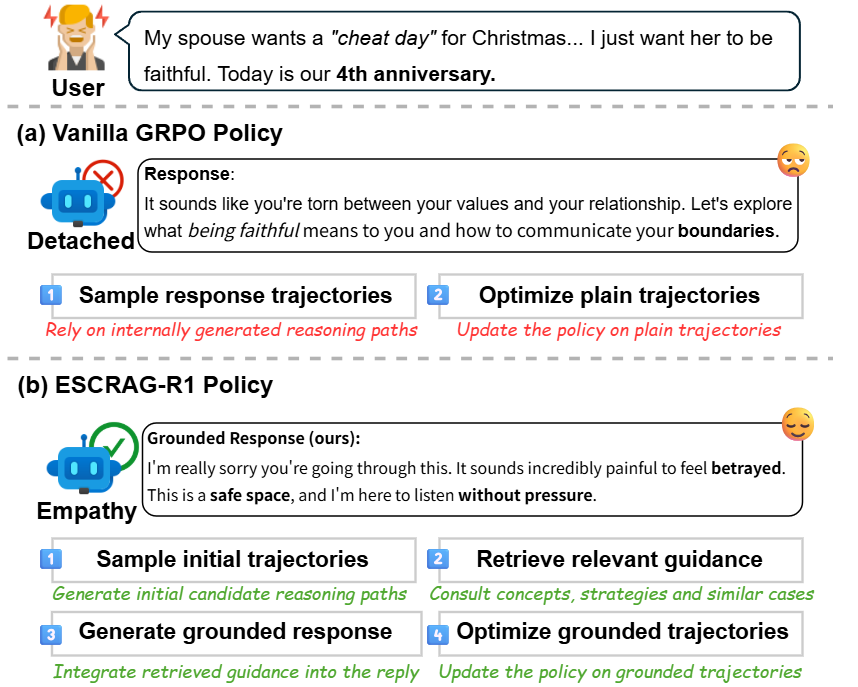}
\caption{Comparison of GRPO and ESCRAG-R1. By integrating retrieved guidance into GRPO, ESCRAG-R1 produces more grounded and empathetic responses.}
\label{fig:banner}
\end{figure}

%% file: chapter/2_Related_Work.tex
\input{picture/main-pic}
\section{Related Work}
\paragraph{Parameter-based ESC Systems.}
Parameter-based approaches internalize supportive behaviors through optimization on specialized datasets. As the foundation for training, studies propose psychological counseling dialogue datasets with diverse topics and strategies to support subsequent model training \citep{zhang2024cpsycoun,liu2021towards,zheng2023building}. Many works adopt supervised fine-tuning (SFT) on counseling dialogues, enabling models to learn therapeutic patterns from the data and thereby deliver better services \citep{chen2023soulchat,zhou2025diacbt}. In reinforcement learning, some studies construct preference pair data and employ Direct Preference Optimization (DPO) \citep{rafailov2023direct} to align models with human preferences \citep{zhao-etal-2025-chain}. With the emergence of Group Relative Policy Optimization (GRPO), several works have conducted preliminary explorations, though these attempts rely entirely on LLMs to explore strategies autonomously, lacking external guidance or structured intervention \citep{wang2025compeer,yang2025towards,wang2025rlver}. 
However, these parameter-based methods mainly learn supportive patterns from fixed training data, without explicitly incorporating practice-grounded counseling knowledge into policy optimization. As a result, they may lack guidance for stage-aware reasoning and context-sensitive intervention, motivating our retrieval-augmented reinforcement learning framework.

\paragraph{Inference-augmented ESC Systems.}
Inference-augmented approaches enhance ESC systems by introducing external knowledge or structured reasoning during inference without modifying model parameters. For example, some studies develop dialogue agent frameworks to construct ESC systems that are more empathetic and helpful \citep{zhang2024escot,zhang-etal-2025-intentionesc,li2024helpful}. Beyond reasoning-focused approaches, a growing body of research draws upon established psychological theories to ground system behavior in clinically validated principles \citep{xiao2024healme,xu2025autocbt}. Additionally, recent studies adopt retrieval-augmented generation (RAG) frameworks to incorporate external knowledge or relevant dialogue exemplars during inference \citep{xiong2024dq,li2024uncertaintyrag,liu2025longemotion}. Despite their strengths, the augmented information in these approaches often remains peripheral to the model's core decision-making, serving as auxiliary input rather than actively shaping the underlying intervention policy. This motivates us to integrate retrieval-based psychological guidance into reinforcement learning, so that external counseling knowledge can participate in policy optimization rather than only assisting inference.

%% file: picture/main-pic.tex
\begin{figure*}[!t]
\setlength\abovecaptionskip{0.2\baselineskip}
\setlength\belowcaptionskip{0.2\baselineskip}
\centering
\includegraphics[width=0.95\textwidth]{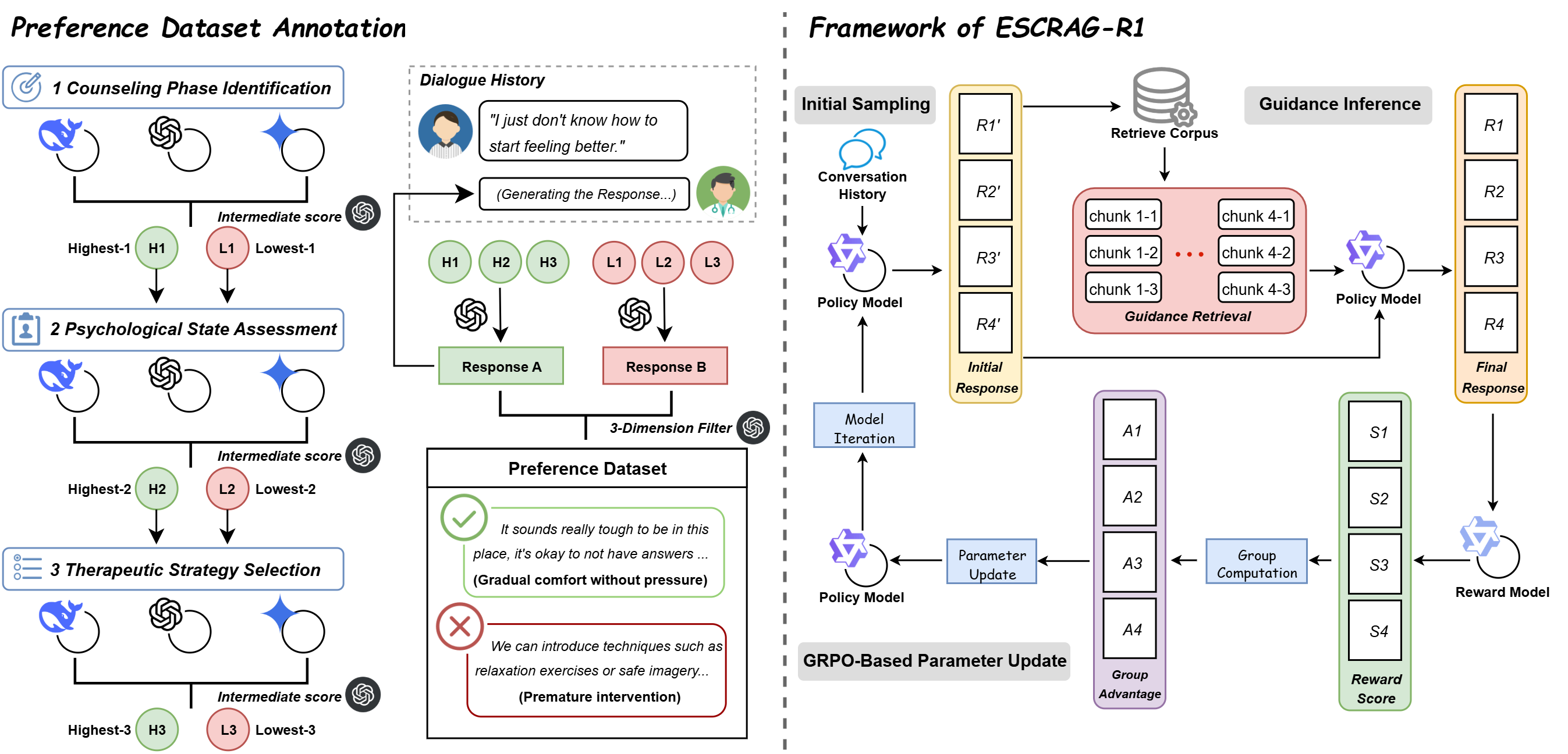}
\caption{The left panel shows the stage-aware annotation pipeline for ESC-Preference, and the right panel illustrates how retrieval-augmented guidance is integrated into GRPO to optimize grounded, empathetic support responses.}
\label{fig:main-pic}
\end{figure*}

%% file: chapter/3_Method.tex
\section{Method}
The overall framework of ESCRAG-R1 consists of two main components: 
\textbf{ESC-Preference Construction} and \textbf{Retrieval-Guided Policy Optimization}, 
as shown in Figure~\ref{fig:main-pic}. 
ESC-Preference is constructed to provide both preference supervision for reward modeling and high-quality counseling exemplars for retrieval grounding. 
Based on this dataset, we first train a multi-perspective reward model, and then optimize the ESC policy through supervised fine-tuning and retrieval-augmented GRPO.

\subsection{Task Definition}

Emotional Support Conversation (ESC) is formulated as a turn-level counselor response generation task. 
A dialogue consists of alternating user and counselor utterances, denoted as 
$\mathcal{D}=\{u_1,c_1,u_2,c_2,\dots,u_T,c_T\}$, where $u_i$ is the $i$-th user utterance, $c_i$ is the corresponding counselor response, and $T$ is the number of counselor turns. 
At the $i$-th counselor turn, the model observes the dialogue history 
$\mathcal{H}_i=\{u_1,c_1,\dots,u_{i-1},c_{i-1},u_i\}$ and generates a response $c_i$ according to the counselor policy $\pi_\theta(\cdot \mid \mathcal{H}_i)$, where $\theta$ denotes the policy parameters.

The goal is to learn a counselor policy $\pi_\theta$ that maximizes the expected reward of generated responses over observed dialogue histories:
\begin{equation}
\label{eq:task-objective}
\pi_\theta^\ast
=
\arg\max_{\pi_\theta}\;
\mathbb{E}_{\mathcal{H}_i,\,
c_i \sim \pi_\theta(\cdot \mid \mathcal{H}_i)}
\left[
R(\mathcal{H}_i, c_i)
\right],
\end{equation}
where $R(\mathcal{H}_i, c_i)$ denotes the reward function that evaluates the emotional support quality of response $c_i$ under dialogue history $\mathcal{H}_i$.
% \subsection{Task Definition}
% Emotional Support Conversation (ESC) is formulated as a turn-level counselor response generation task. At each counselor turn, the model generates an emotionally supportive response conditioned on the preceding dialogue context. A dialogue consists of alternating user and counselor utterances, denoted as 
% $\mathcal{D}^{(n)} = \{u_1, c_1, u_2, c_2, \dots, u_T, c_T\}$\yuxuan{All symbols should be explicitly defined, in particular $u$ and $c$. Moreover, the notation ${D}^{(n)}$ is inconsistent, as it appears to have been replaced by $u^T$.}. 
% At counselor turn $i$, the dialogue state is defined as the history up to the current user utterance, 
% $\mathcal{H}_i = \{u_1, c_1, \dots, u_i\}$, 
% and the model generates the corresponding response 
% $c_i \sim \pi_\theta(\cdot \mid \mathcal{H}_i)$, 
% where $\pi_\theta$ denotes the counselor policy.\yuxuan{Replacing this part with an objective function would make the formulation clearer.}

\subsection{ESC-Preference Construction}
\label{subsec:dataset-annotation}
To obtain reward signals that balance professional competence and empathetic quality, as well as practice-grounded guidance for both training and inference, we construct the ESC-Preference dataset, which provides multi-perspective preference pairs and high-quality counseling exemplars. Based on ESConv~\citep{liu2021towards}, we simulate multi-turn client--counselor interactions to construct counseling dialogues. As shown in Figure~\ref{fig:main-pic}, candidate models perform three-stage counseling inference to build contrasting reasoning paths, from which reliable preference pairs are obtained through stage-wise selection and final preference filtering. The prompts are provided in Appendix~\ref{sec:dataset-construction-appendix}.

% \subsection{ESC-Preference Construction}
% \yuxuan{motivation: why need the dataset,}
% \label{subsec:dataset-annotation}
% We construct the ESC-Preference dataset to support reward model training and retrieval augmentation. The dataset is generated through an interactive self-play framework combined with a stage-wise reasoning exploration process. At each counselor turn, two contrasting reasoning trajectories are constructed to produce candidate responses, which are subsequently evaluated under a multi-dimensional framework to obtain reliable preference pairs. The prompts and the detailed counseling framework during construction can be seen in Appendix \ref{sec:dataset-construction-appendix}.\yuxuan{It is recommended to briefly describe the process by following the subsubsection titles.}\yuxuan{ref figure 2}

\paragraph{Stage-wise Response Reasoning.}
Given a dialogue history, we construct candidate responses through a three-stage counseling reasoning framework. This design follows the progressive nature of counseling interactions, where the counselor first understands the counseling context, then assesses the client's psychological state, and finally selects an intervention strategy. This process grounds response generation in counseling-oriented reasoning rather than relying only on surface dialogue context. To obtain contrasting response candidates, three candidate models, including GPT-4o~\citep{openai-gpt-4o}, Gemini-2.5-Pro~\citep{comanici2025gemini}, and DeepSeek-V3~\citep{deepseekai2024deepseekv3technicalreport}, independently perform the three-stage inference. At each stage, intermediate selection is applied to retain both high- and low-quality reasoning states. These selected states are propagated through the subsequent stages, producing two contrasting reasoning trajectories, which are then used by GPT-4o to generate paired counselor responses.

% \paragraph{Self-Play Dialogue Generation.}
% \yuxuan{motivation: Motivation for this dataset should be clarified, preferably with an analysis of the limitations of existing methods.}We adopt an interactive self-play paradigm to generate emotionally grounded counseling dialogues.\yuxuan{the simple present tense} Specifically, the problem descriptions in the ESConv \citep{liu2021towards} are used as initial psychological profiles for the client. A GPT-4o \citep{openai-gpt-4o} client simulator is instructed to consistently role-play the client according to the given profile, including emotional state, experiences, and situational context. Starting from this persona, the simulator engages in multi-turn interactions with the counselor model, producing dialogue trajectories that simulate realistic emotional support conversations. The response generated from the highest-scoring reasoning trajectory is appended to the dialogue history and used to continue the interaction with the client simulator.
\paragraph{Multi-perspective Preference Evaluation.}
To evaluate the paired responses beyond a single overall score, we use GPT-5-chat~\citep{openai2025gpt5} as the evaluator and adopt a multi-perspective evaluation framework covering the client, counselor, and judge viewpoints. The client perspective focuses on perceived emotional support quality, the counselor perspective evaluates therapeutic coherence and professional grounding, and the judge perspective assesses whether the response fits the client's state and targets the core problem. Based on these evaluation results, we apply a dominance-based filtering criterion: the preferred response must achieve non-negative improvement across all metrics and strictly outperform the alternative response in at least three dimensions. Only pairs satisfying these conditions are retained. Through this process, we obtain 3,667 high-confidence preference pairs, containing 7,334 responses in total.

\subsection{Retrieval-Guided Policy Optimization}
\label{subsec:retrieval-guided-policy-optimization}

To move beyond inference-only guidance and enable retrieved counseling knowledge to shape the model policy, we use ESC-Preference to guide policy optimization. 
The preference pairs are used to train a reward model, and the preferred responses are organized as a retrieval corpus to provide counseling guidance. 
We first initialize the policy with a supervised reasoning-response pattern, and then integrate retrieval augmentation into GRPO. 
In this way, retrieved counseling exemplars are used during both policy optimization and inference, rather than serving only as inference-time guidance.

\paragraph{Preference-Guided Reward Learning.}
To provide reward signals for policy optimization, we train a reward model on the preference pairs in ESC-Preference. Each preference pair is represented as $(\mathcal{H}_i, c_i^{+}, c_i^{-})$, where $c_i^{+}$ and $c_i^{-}$ denote the preferred and rejected responses under the same dialogue state $\mathcal{H}_i$, respectively. The reward model estimates the relative quality of candidate responses by mapping each dialogue state and response to a scalar reward score. Specifically, it consists of a pretrained backbone $f_\theta(\cdot)$ and a linear scoring head $g_\psi(\cdot)$, and the reward score $r_i^{\mathrm{RM}}$ for a candidate response is computed as:
\begin{equation}
r_i^{\mathrm{RM}} = g_\psi\!\left(f_\theta(\mathcal{H}_i, c_i)\right).
\end{equation}

During training, the backbone parameters $\theta$ are frozen and only the
scoring head parameters $\psi$ are optimized. For each preference pair, the reward model produces scores
$r_i^{RM+} = R_\phi(\mathcal{H}_i, c_i^{+})$ and
$r_i^{RM-} = R_\phi(\mathcal{H}_i, c_i^{-})$
 for the preferred and rejected responses respectively. The loss function of the reward model is defined as:
\begin{equation}
\begin{aligned}
\mathcal{L}_{RM}
= \frac{1}{N}\sum_{i=1}^{N}
\Big[
-\log \sigma(r_i^{RM+} - r_i^{RM-}) \\
+ \lambda (r_i^{RM+} + r_i^{RM-})^2
\Big],
\end{aligned}
\end{equation}
where $\sigma(\cdot)$ denotes the sigmoid function and
$\lambda$ is a regularization coefficient.
The first term encourages the reward model to assign higher scores to
preferred responses, while the second term constrains the reward scale
to prevent score drifting during training. 

During policy optimization, we further introduce an auxiliary
\textit{format reward} to encourage the policy model to follow the structured
output format. Instead of directly generating the final counselor response, the model is expected to first organize its counseling reasoning and then provide the final supportive response. The format reward $r_i^{format}$ is defined as:
\begin{equation}
r_i^{format} =
\begin{cases}
1, & \text{if format satisfied}; \\
0, & \text{otherwise}.
\end{cases}
\end{equation}

The final reward $r_i$ used during
reinforcement learning is defined as:

\begin{equation}
r_i = r_i^{RM} + r_i^{format}.
\end{equation}

\paragraph{Supervised Pattern Initialization.}
Before reinforcement learning, we initialize the policy model using a
supervised fine-tuning (SFT) stage to establish a structured reasoning
pattern for emotional support responses. Specifically, we generate
500 cold-start demonstrations using GPT-4o \citep{openai-gpt-4o}.
Following the prompt design in Appendix~\ref{sec:appendix-policy-model-prompt}, each demonstration consists of a reasoning segment and a final response, formatted as \texttt{<think></think>} and \texttt{<response></response>}.

Given a dialogue state $\mathcal{H}_i$, let $y_i$ denote the target structured output in the cold-start demonstration, which contains both the reasoning segment and the final response. The policy model $\pi_\theta$ is trained to generate $y_i$ conditioned on $\mathcal{H}_i$. The supervised fine-tuning objective is defined as:
\begin{equation}
\mathcal{L}_{\text{SFT}}
= - \mathbb{E}_{(\mathcal{H}_i, y_i)}
\left[
\log \pi_\theta(y_i \mid \mathcal{H}_i)
\right].
\end{equation}

This stage enables the model to learn the reasoning–response
output format and provides a stable initialization for subsequent
reinforcement learning with GRPO.

\paragraph{Retrieval-Augmented GRPO.}
After supervised pattern initialization, we further optimize the counselor policy using GRPO with retrieval augmentation. Different from inference-only RAG, our goal is to expose the policy to retrieved counseling exemplars during the rollout process, so that retrieval-based guidance can influence both response generation and policy updates. Given a dialogue state $\mathcal{H}_i$, the policy first generates an initial response $\hat{c}_i$:
\begin{equation}
\hat{c}_i \sim \pi_\theta(\cdot \mid \mathcal{H}_i).
\end{equation}

The initial response is used as a query to retrieve relevant counseling exemplars from the retrieval corpus constructed during dataset generation. Let $\mathcal{E}_i = \{e_1,\dots,e_k\}$ denote the retrieved exemplars. The policy then generates the final response conditioned on the dialogue state, the initial response, and the retrieved exemplars:
\begin{equation}
c_i \sim \pi_\theta(\cdot \mid \mathcal{H}_i,\hat{c}_i,\mathcal{E}_i).
\end{equation}

\input{table/main-experiment}
During training, for each dialogue state $\mathcal{H}_i$, the current policy samples a group of $N$ retrieval-augmented candidate responses $\{c_i^j\}_{j=1}^{N}$, and each response is evaluated using the reward signal defined above. The group-relative advantage $A_i^j$ for each sampled response is computed as:
\begin{equation}
A_i^j =
\frac{
r_i^j - \frac{1}{N}\sum_{k=1}^{N} r_i^k
}{
\sqrt{
\frac{1}{N}\sum_{k=1}^{N}
\left(
r_i^k-\frac{1}{N}\sum_{l=1}^{N} r_i^l
\right)^2
+\epsilon
}
},
\end{equation}
where $r_i^j$ denotes the reward of the $j$-th sampled response, while $r_i^k$ is used as the summation term over all responses in the same group. The constant $\epsilon$ is used for numerical stability.

We define the importance sampling ratio $\rho_i^j$ between the current policy and the old policy for the $j$-th sampled response as:
\begin{equation}
\rho_i^j =
\frac{\pi_\theta(c_i^j \mid \mathcal{H}_i,\hat{c}_i,\mathcal{E}_i)}
{\pi_{\theta_{\mathrm{old}}}(c_i^j \mid \mathcal{H}_i,\hat{c}_i,\mathcal{E}_i)},
\end{equation}
where $\rho_i^j$ measures how the likelihood of the sampled response changes under the current policy $\pi_\theta$ compared with the old policy $\pi_{\theta_{\mathrm{old}}}$.

Based on the importance sampling ratio $\rho_i^j$ and the group-relative advantage $A_i^j$, the policy is optimized with the following clipped GRPO objective:
\begin{equation}
\begin{aligned}
\mathcal{L}_{\mathrm{GRPO}}(\theta)
&=
-\mathbb{E}_{\mathcal{H}_i,\{c_i^j\}_{j=1}^{N}}
\Bigg[
\frac{1}{N}\sum_{j=1}^{N}
\\
&\hspace{-1.4cm}
\min \Big(
\rho_i^j A_i^j,\,
\mathrm{clip}(\rho_i^j,1-\delta,1+\delta)A_i^j
\Big)
\Bigg],
\end{aligned}
\end{equation}
where $\mathrm{clip}(\rho_i^j,1-\delta,1+\delta)$ constrains the policy ratio within $[1-\delta,1+\delta]$, and $\delta$ is a clipping coefficient that limits excessively large policy updates. The overall procedure of retrieval-augmented GRPO and RAG inference is summarized in Appendix ~\ref{subsec:ragrpo-and-rag-inference}.

%% file: table/main-experiment.tex
\begin{table*}[!t]
\setlength\abovecaptionskip{0.2\baselineskip}
\setlength\belowcaptionskip{0.2\baselineskip}
\centering
\begin{tabular}{l
>{\centering\arraybackslash}p{0.6cm}  % EI列
>{\centering\arraybackslash}p{0.6cm}  % EL列
>{\centering\arraybackslash}p{0.6cm}  % TA列
>{\centering\arraybackslash}p{0.6cm}  % SC列
>{\centering\arraybackslash}p{0.6cm}  % TH列
>{\centering\arraybackslash}p{0.6cm}  % SF列
>{\centering\arraybackslash}p{0.6cm}  % PT列
c}  % Avg列
\toprule
\multirow{2}{*}{\textbf{Model}} 
& \multicolumn{3}{c}{\textbf{Client}}                    
& \multicolumn{2}{c}{\textbf{Counselor}}
& \multicolumn{2}{c}{\textbf{Judge}}
& \multirow{2}{*}{\textbf{Avg}} \\

\cmidrule(lr){2-4}
\cmidrule(lr){5-6}
\cmidrule(lr){7-8}

& EI & NE & TA 
& SC & TH 
& SF & PT 
&  \\

\midrule

\textit{General-Purpose Models (Raw)}\\
GPT-4o    & 8.19 & 8.61 & 8.07 & 8.87 & 7.90 & 8.40 & 7.56 & 8.23  \\
GPT-4o \textit{with RAG} & \textbf{8.79} & 8.96 & 8.56 & \underline{8.95} & \textbf{8.13} & \textbf{8.76} & \textbf{7.96} & \underline{8.59}\\
Llama-3.1-8B-Instruct & 8.20 & 8.48 & 8.06 & 8.71 & 7.78 & 8.27 & \underline{7.65} & 8.16 \\
Llama-3.1-8B-Instruct \textit{with RAG} & 8.73 & 8.50 & \underline{8.59} & 8.09 & 7.71  & 8.29 & 7.57 & 8.21 \\
% Gemini-2.5-pro   & 8.94 & 9.46 & 8.75 & 9.40  & 8.51 & 9.12 & 8.50  & 8.95 \\
% Deepseek-V3 & 8.68 & 9.13 & 8.40 & 8.99 & 8.13 & 8.86 & 8.04 & 8.60 \\
% Deepseek-V3 \textit{with RAG} & 8.97 & 9.41 & 8.79 & 8.85 & 8.21 & 9.00 & 8.23 & 8.78 \\

% Qwen-2.5-3B-Instruct & 7.18& 7.20 & 7.18 & 7.97 & 6.92  & 7.08 & 6.31 & 7.12 \\
% Qwen-2.5-3B-Instruct \textit{with RAG} & 7.93 & 8.16 & 7.81 & 7.18 & 6.90 &  7.00& 6.12 & 7.30\\
Qwen-2.5-7B-Instruct & 7.62 & 7.63 & 7.51 & 8.49 & 7.36 & 7.63 & 6.92 & 7.59 \\
Qwen-2.5-7B-Instruct \textit{with RAG} & 8.31 & 8.50 & 8.12 & 8.22 & 7.49 & 8.00 &  7.18 & 7.97\\
\midrule
\textit{Specialized ESC models} \\
ChatCounselor-7B & 6.22 & 5.59 & 6.25 & 7.06 & 6.07& 6.09 & 5.56 & 6.12 \\
PsyLLM-8B & 8.59 & 9.18 & 8.34 & 8.68 & 7.87 & 8.55 & \underline{7.65} & 8.41 \\

\midrule
\textit{ESCRAG-R1}
\\
ESCRAG-R1-3B \textit{with RAG}& 8.68 & \underline{9.31} & 8.55 & 8.80  & 8.02 & 8.58 & 7.56 & 8.50 \\
ESCRAG-R1-7B \textit{with RAG}& \underline{8.74} & \textbf{9.60} & \textbf{8.68} & \textbf{8.98} & \underline{8.09} & \underline{8.66} & \underline{7.65} & \textbf{8.63} \\

\bottomrule
\end{tabular}
\caption{
Evaluation results from three perspectives. 
Client: EI (Emotional Impact), NE (Non-Lecturing Empathy), TA (Therapeutic Alliance); 
Counselor: SC (Strategic Coherence), TH (Theoretical Application); 
Judge: SF (Strategy-Client Fit), PT (Problem Targeting).
The best and second-best results are \textbf{bolded} and \underline{underlined}, respectively.
}
\label{tab:main_results}
\end{table*}

%% file: chapter/4_Experiment.tex
\section{Experiment}
\label{sec-experiment}

\subsection{Experiment Settings}
\label{subsec-experiment-settings}
\paragraph{Dataset.}
We conduct experiments on ESConv~\citep{liu2021towards} using its default training and test split. The training set contains 910 dialogues with 10,939 turns, which are used for reward model training, retrieval corpus construction, and GRPO-based policy optimization. During evaluation, we assess model performance on the test split, which contains 195 dialogues with a total of 2,505 turns.

\paragraph{Evaluation Metrics.}

We evaluate model performance from three complementary perspectives: Client, Counselor, and Judge.  For automatic evaluation, we use GPT-5-chat as the evaluator under a unified scoring prompt. Detailed descriptions of the metrics are provided in Appendix \ref{subsec:evaluator}.

\paragraph{Implementation Details.}
For reward model, we adopt Qwen-2.5-7B \citep{qwen2,qwen2.5}  as the backbone. For policy optimization, we use Qwen-2.5-3B-Instruct and Qwen-2.5-7B-Instruct as the backbone policy models. For the retrieval component, we use bge-m3 \citep{bge-m3} to encode dialogue chunks and construct the vector database. For fairness, all RAG-based experiments use the same retrieval number, with $N=3$ retrieved exemplars. Detailed hyperparameter settings are provided in Appendix \ref{appendix-parameter-setting}.

\input{table/ablation-exp}

\paragraph{Baselines.}
We compare our method against two categories of baselines: general-purpose large language models and specialized emotional support dialogue models. The general-purpose models include GPT-4o~\citep{openai-gpt-4o}, Llama-3.1-8B-Instruct \citep{grattafiori2024llama3herdmodels}, and Qwen-2.5-7B-Instruct\citep{qwen2,qwen2.5}, while the specialized models include ChatCounselor-7B \citep{liu2023chatcounselor} and PsyLLM-8B \citep{hu2025beyond}. Considering the different architectures and usage settings of the models, we additionally evaluate the RAG setting only for general-purpose models.

\subsection{Main Results}
\label{subsen-experiment-results}

Table~\ref{tab:main_results} presents the overall comparison results from the Client, Counselor, and Judge perspectives. 
The first key finding is that ESCRAG-R1 achieves the strongest overall performance among all compared models. 
Specifically, ESCRAG-R1-7B with RAG obtains the highest average score of 8.63, while ESCRAG-R1-3B with RAG also achieves a competitive score of 8.50. 
These results demonstrate that the proposed framework is effective across different model scales and consistently improves the overall quality of ESC responses.

The second key finding is that ESCRAG-R1 achieves a better balance between empathetic expression and professional grounding. 
Compared with GPT-4o with RAG, ESCRAG-R1-7B obtains a substantially higher score on Non-Lecturing Empathy (9.60 vs. 8.96) and a higher score on Therapeutic Alliance (8.68 vs. 8.56), indicating that our model produces responses that are perceived as more supportive and less didactic even when compared with a strong retrieval-augmented general-purpose model. 
Meanwhile, ESCRAG-R1-7B also maintains strong Counselor-side performance, with 8.98 on Strategic Coherence and 8.09 on Theoretical Application. 
These results show that ESCRAG-R1 improves empathetic expression while preserving professional therapeutic reasoning, thereby better supporting empathy-expertise alignment.

The third key finding is that specialized ESC models still show limitations in jointly maintaining empathy and professional grounding. 
Although PsyLLM-8B achieves a strong average score of 8.41 and performs well on empathy-related metrics, ESCRAG-R1-7B further improves all three Client-side metrics and achieves higher Counselor-side scores, increasing Strategic Coherence from 8.68 to 8.98 and Theoretical Application from 7.87 to 8.09. 
Meanwhile, ChatCounselor-7B obtains a much lower average score of 6.12, with weak performance across both Client and Judge dimensions. 
These results suggest that counseling-oriented training alone may be insufficient to consistently produce responses that are both empathetic and strategically appropriate, highlighting the importance of integrating structured reasoning and preference-based policy optimization.
\input{table/case_study}
\subsection{Ablation Analysis}
\label{sub-sec:ablation}
Table~\ref{tab:ablation_3b} reports the ablation study on the core components of the framework, examining how each module contributes to the overall performance. 
Figure~\ref{fig:retrieval_number} further investigates how varying the number of retrieved exemplars affects model performance.

\paragraph{Ablation on Framework Components.}
Table~\ref{tab:ablation_3b} shows that each component of ESCRAG-R1 contributes to the overall performance. 
Starting from the backbone, Qwen-2.5-3B-Instruct achieves an average score of 7.12, and adding RAG brings a modest improvement to 7.30, suggesting that retrieved exemplars provide useful supportive cues but are insufficient when used only as external inference-time context. 
After Cold-Start SFT establishes a structured reasoning-and-response pattern, \textit{Vanilla-GRPO} further improves the average score from 8.14 to 8.35, showing that reinforcement learning can enhance the general quality of emotional support generation. 
Adding \textit{RAG Inference} on top of Vanilla-GRPO further raises the score to 8.44, indicating that retrieved examples can provide additional guidance during generation. 
The full ESCRAG-R1-3B framework achieves the best overall score of 8.50, outperforming both \textit{Vanilla-GRPO} and \textit{Vanilla-GRPO + RAG Inference}. 
Notably, compared with \textit{Vanilla-GRPO + RAG Inference}, ESCRAG-R1 further improves the Client-side metrics such as Emotional Impact and Therapeutic Alliance, as well as the Judge-side metrics of Strategy--Client Fit and Problem Targeting. 
These improvements suggest that integrating retrieval into policy optimization helps the model better internalize practice-grounded counseling strategies, leading to more supportive responses that are better aligned with the client's state and core problem.

\input{picture/retrieval_number}
\paragraph{Ablation on Retrieval Number.}
Figure~\ref{fig:retrieval_number} illustrates the effect of different retrieval numbers on ESCRAG-R1-7B. Overall, increasing the retrieval number from \(n=2\) to \(n=3\) leads to consistent improvements across all evaluation dimensions, raising the average score from 8.44 to 8.63. This suggests that using too few retrieved exemplars may provide insufficient counseling guidance for response refinement. When the retrieval number further increases to \(n=4\), the overall performance remains competitive but slightly decreases to 8.61, indicating that additional retrieved examples do not necessarily bring further gains and may introduce redundant or less focused guidance.

% The second group of results in Table~\ref{tab:ablation_3b} examines the effect of varying the retrieval number \(n\), which denotes the number of retrieved examples used in both training and inference. The results show that performance becomes strong at \(n=3\), suggesting that a moderate retrieval size is sufficient to provide informative and relevant supportive guidance. At this setting, the model achieves the best results on all three Client-side dimensions, namely Emotional Impact (EI), Empathy vs. Lecturing (EL), and Therapeutic Alliance (TA), as well as on both Judge-side dimensions, Strategy-Client Fit (SF) and Problem Targeting (PT). This indicates that \(n=3\) is particularly effective for generating responses that are emotionally supportive and well aligned with the client’s needs. In contrast, \(n=4\) yields the best performance on the two Counselor-side dimensions, Strategic Coherence (SC) and Theoretical Application (TH), suggesting that a larger retrieval set benefits response structure and theoretical grounding. Overall, these results reveal a trade-off between client-oriented emotional quality and counselor-oriented reasoning quality.
\input{picture/human-evaluation}
\subsection{Human Evaluation}
\label{human-eval}
We further conduct human evaluation on 100 randomly sampled dialogue contexts from the ESConv test set. 
As shown in Figure~\ref{fig:human_eval}, ESCRAG-R1-7B with RAG consistently achieves more wins than losses against all three baselines, obtaining 57 wins against GPT-4o with RAG, 54 wins against PsyLLM-8B without RAG, and 63 wins against Llama-3.1-8B-Instruct with RAG. 
Interestingly, although GPT-4o with RAG performs strongly in LLM-as-Judge evaluation, human annotators still prefer ESCRAG-R1 in more cases, and the comparison with PsyLLM-8B is relatively close. 
This suggests that models whose policies are updated with psychological counseling data may produce responses that better match human expectations.

\subsection{Case Study}
Table~\ref{tab:case_study} illustrates how retrieval augmentation helps ESCRAG-R1 refine its response. 
Compared with the initial response, the final response provides warmer emotional validation, avoids premature advice, and maintains a more supportive stance. 
The retrieved chunks guide the model to acknowledge the client's hurt and encourage further exploration rather than directly prescribing solutions. 
Since retrieval-guided responses are evaluated during GRPO training, such guidance can further shape the policy beyond inference-time generation.

%% file: table/ablation-exp.tex
\begin{table*}[!t]
\setlength\abovecaptionskip{0.2\baselineskip}
\setlength\belowcaptionskip{0.2\baselineskip}
\centering
\begin{tabular}{lcccccccc}
\toprule
\multirow{2}{*}{\textbf{Model}} 
& \multicolumn{3}{c}{\textbf{Client}}                    
& \multicolumn{2}{c}{\textbf{Counselor}}
& \multicolumn{2}{c}{\textbf{Judge}}
& \multirow{2}{*}{\textbf{Avg}} \\

\cmidrule(lr){2-4}
\cmidrule(lr){5-6}
\cmidrule(lr){7-8}

& EI & NE & TA 
& SC & TH 
& SF & PT 
&  \\
\midrule
\textit{Ablation on Framework Components} \\
Qwen-2.5-3B-Instruct & 7.18& 7.20 & 7.18 & 7.97 & 6.92  & 7.08 & 6.31 & 7.12 \\
Qwen-2.5-3B-Instruct \textit{with RAG} & 7.93 & 8.16 & 7.81 & 7.18 & 6.90 &  7.00& 6.12 & 7.30 $\uparrow0.18$\\
Cold-Start SFT
& 8.14 & 8.46 & 8.00 & 8.88 & 7.88 & 8.21 & 7.43 & 8.14 $\uparrow1.02$\\

\textit{+ Vanilla-GRPO}
& 8.19 & 9.06 & 8.13 & 9.00 & 8.02 & 8.52 & 7.55 & 8.35 $\uparrow1.23$\\

\textit{+ Vanilla-GRPO + RAG Inference }
& 8.49 & 9.30 & 8.35 & 8.96 & 8.03 & 8.46 & 7.52 & 8.44 $\uparrow1.32$\\

ESCRAG-R1-3B \textit{+ RAG Inference}
& 8.68 & 9.31 & 8.55 & 8.80 & 8.02 & 8.58 & 7.56 & 8.50 $\uparrow1.38$\\
\midrule

\end{tabular}
\caption{Ablation on framework components. Vanilla-GRPO uses standard GRPO training, RAG Inference applies retrieval augmentation only at test time, and ESCRAG-R1-3B uses our full joint training framework with retrieval augmentation during both training and test time.}
\label{tab:ablation_3b}
\end{table*}

%% file: table/case_study.tex
\begin{table*}[t]
\small
\centering
\renewcommand{\arraystretch}{1.15}
\setlength{\tabcolsep}{6pt}

\begin{tabularx}{\textwidth}{
>{\centering\arraybackslash}m{0.18\textwidth}
>{\RaggedRight\arraybackslash}m{0.35\textwidth}
>{\RaggedRight\arraybackslash}m{0.39\textwidth}
}
\toprule

\multicolumn{3}{>{\RaggedRight\arraybackslash}p{0.96\textwidth}}{
\textbf{Client:} I feel like my friend group judges me for simply being myself. Although they are probably joking, I sometimes take their words personally, and it makes me upset.

\textbf{Counselor:} It sounds like your friends' jokes can feel judgmental to you, and even if they do not mean harm, their words still end up hurting you.

\textbf{Client:} Should I still hang out with them if they keep making me feel this way?
} \\

\midrule

\textbf{Response}
&
\textbf{Initial Response (Vanilla-GRPO)} \par\vspace{2pt}
It sounds like this situation is really tough for you. It's understandable that you feel upset when people you care about are not being supportive. \newline
Maybe we can explore together what being in this situation feels like for you and what boundaries, if any, you might want to set. Would you be comfortable discussing how you want to approach your friend group moving forward?
&
\textbf{Final Response (ESCRAG-R1)} \par\vspace{2pt}
It sounds like this situation is really affecting you, and 
\textcolor{highlightorange}{I want to acknowledge how difficult that must be}. 
It makes sense that you feel upset when people you care about say things that hurt, even if they are joking. \par\vspace{2pt}

Would you like to talk more about 
\textcolor{highlightorange}{how this is affecting you}? 
We can take our time to understand your feelings and what matters most to you. 
Remember, \textcolor{highlightorange}{you're not alone in this}, and 
\textcolor{highlightorange}{I'm here to support you} as you explore your options.
\\

\midrule

\raisebox{-3.5em}{\textbf{Retrieved Chunks}}
&
\multicolumn{2}{>{\RaggedRight\arraybackslash}p{0.76\textwidth}}{
\textbf{[Chunk 1]}
\textit{History:} The client is hurt and confused after a friend said she hated them while they were also dealing with family problems. \newline
\textit{Response:} The counselor validates the client's pain and confusion, encourages them to process feelings at their own pace, and reassures them that they do not have to go through this alone. 
\par\vspace{2pt}
\textbf{[Chunk 2]} 
\textit{History:} The client feels rejected because friends are cold and unwilling to meet, then asks what they should do. \newline
\textit{Response:} The counselor acknowledges the hurt, avoids direct advice, invites the client to explore how the coldness affects them, and supports taking things one step at a time.
} \\

\bottomrule
\end{tabularx}

\caption{
Case study of ESCRAG-R1. Retrieved chunks guide the model to produce a more supportive final response, with highlighted text showing improved counseling behaviors. During training, the reward of such retrieval-guided responses is used to update the policy, allowing external guidance to shape model parameters.
}
\label{tab:case_study}
\end{table*}

%% file: picture/retrieval_number.tex
\begin{figure}[htbp]
\setlength\abovecaptionskip{0.2\baselineskip}
\setlength\belowcaptionskip{0.2\baselineskip}
    \centering
    \includegraphics[width=0.95\linewidth]{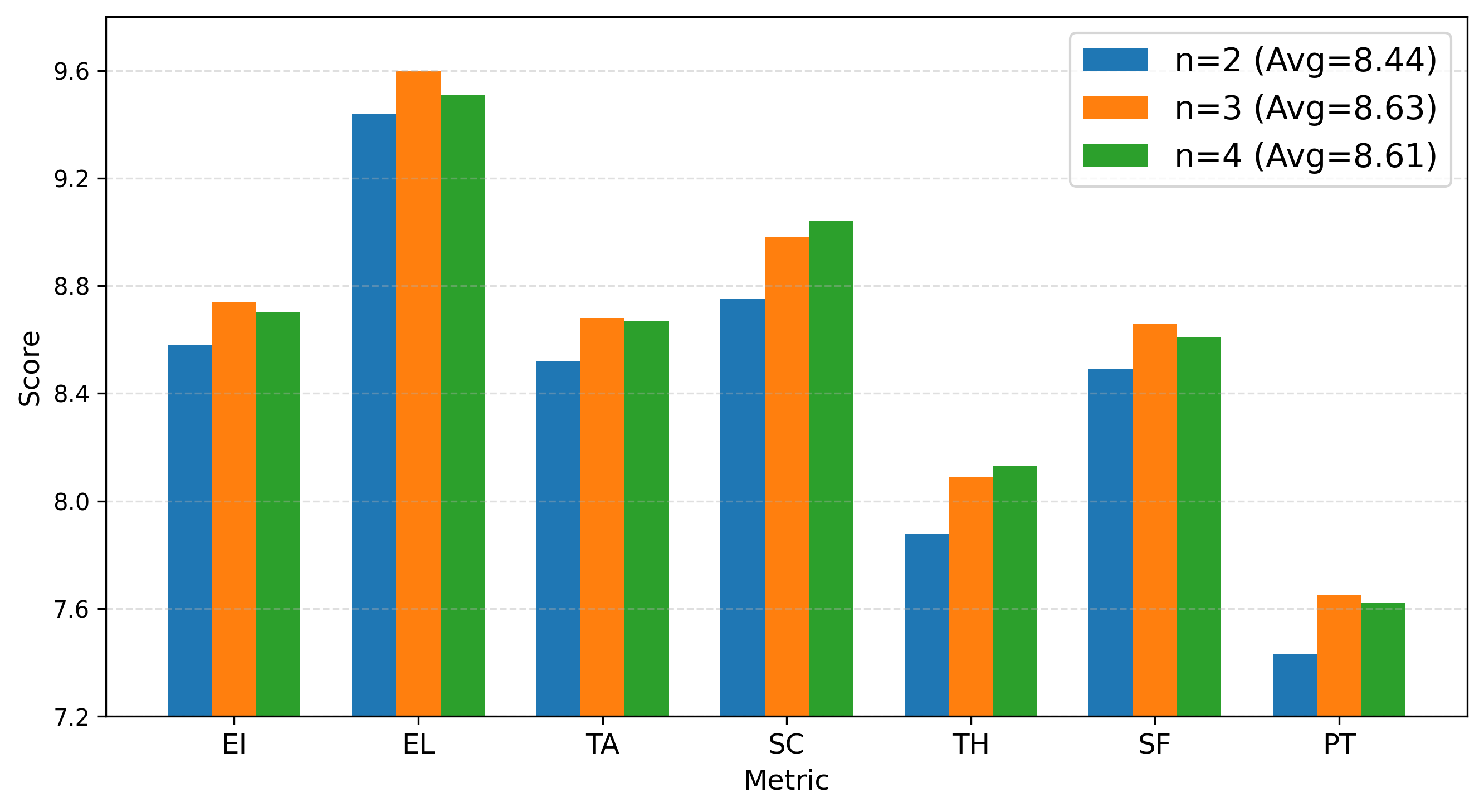}
    \caption{Performance of ESCRAG-R1-7B with different retrieval numbers across evaluation dimensions.}
    \label{fig:retrieval_number}
\end{figure}
\vspace{-5pt}

%% file: picture/human-evaluation.tex
\begin{figure}[htbp]
\setlength\abovecaptionskip{0.2\baselineskip}
\setlength\belowcaptionskip{0.2\baselineskip}
    \centering
    \includegraphics[width=0.90\linewidth]{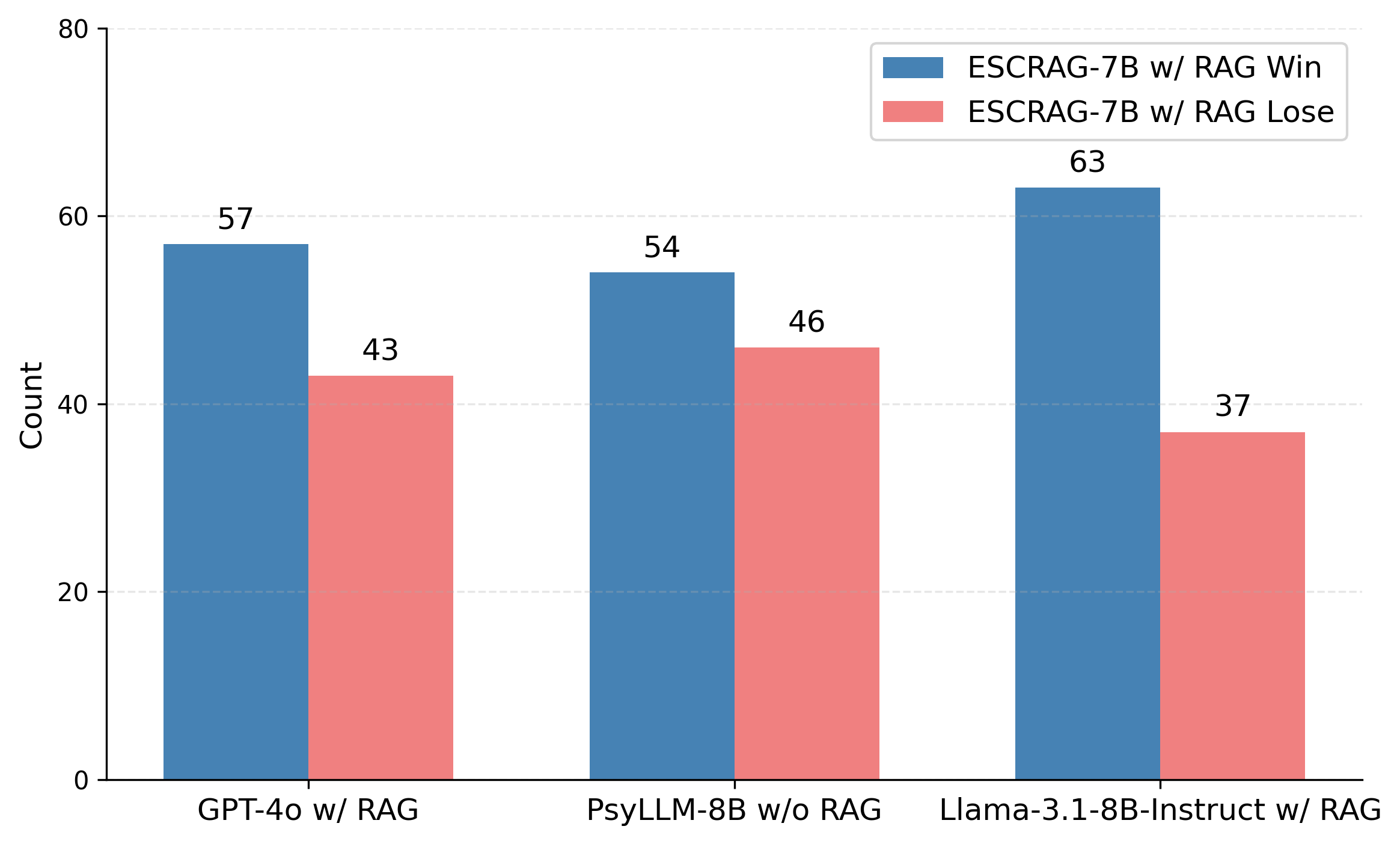}
    \caption{Human evaluation results on 100 sampled dialogue contexts from the ESConv test set.}
    \label{fig:human_eval}
\end{figure}

%% file: chapter/5_Conclusion.tex
\section{Conclusion}
In this paper, we propose \textsc{\textbf{ESCRAG-R1}}, a retrieval-augmented reinforcement learning framework for emotional support conversation. 
We construct \textsc{\textbf{ESC-Preference}} with a multi-perspective evaluation framework to support reward modeling and retrieval grounding. 
Experimental results show that ESCRAG-R1 improves emotional support quality across model scales and achieves stronger empathy-expertise alignment than general-purpose and specialized ESC models.

%% file: chapter/Limitations_Statement.tex
\section{Limitations}

Despite the effectiveness of ESCRAG-R1, this work still has several limitations. 
First, due to computational resource constraints, our experiments are mainly conducted on 3B and 7B policy models. 
Although the results demonstrate consistent improvements across different model scales, we have not fully investigated how the proposed framework performs on larger-scale LLMs. 
Second, this work focuses on text-only Emotional Support Conversation. 
However, real-world emotional support often involves multimodal signals, such as facial expressions, speech tone, and other behavioral cues, which may provide important information about the user's emotional state. 

\section{Ethical Considerations}
This work aims to improve the performance of Emotional Support Conversation (ESC) models and is intended for research purposes only. 
All data and models used in this work are based on publicly available and open-source resources, and our experiments do not involve private user data or personally identifiable information. 

During dataset construction, we carefully design the annotation and generation pipeline to reduce the risk of producing harmful or inappropriate content. 
The prompts used for client simulation, counselor reasoning, and response evaluation are designed to guide the models toward supportive, non-harmful, and therapeutically appropriate interactions, as detailed in Appendix~\ref{sec:dataset-construction-appendix}. 
Nevertheless, we acknowledge that automatically generated emotional support responses may still carry potential risks if used in real-world settings without proper safeguards.

Therefore, ESCRAG-R1 and the constructed dataset are intended only for academic research. 
They should not be directly used as a substitute for professional psychological counselors or mental health services. 
Any future practical deployment should include strict safety mechanisms, human oversight, and clear instructions for users to seek professional help in high-risk situations.

\section{Acknowledgements}
This work was supported in part by Guangdong Province Science and Technology Foundation (No. 2026A1515011806, No. 2024TQ08X559), in part by Innovation Team Project of Guangdong Province (No. 2024KCXTD017), in part by Shenzhen Science and Technology Foundation (No. JCYJ20240813145816022).

%% file: chapter/appendix.tex
\section{Prompt Design for ESC-Preference Construction}
\label{sec:dataset-construction-appendix}
% To construct high-quality preference pairs for emotional support conversation, we design a multi-role framework involving three collaborative agents: Client, Counselor and Evaluator. These roles interact through a controlled pipeline to produce reliable and diverse preference data.

\subsection{Client Simulator Prompt}
We design a role-playing prompt to simulate the client as a patient seeking emotional support. The prompt enforces the client to remain consistent with the given emotional and situational context, respond naturally to the therapist, and generate only one short and succinct sentence for the next turn without additional explanation. The prompt design can be seen in Figure \ref{fig:client_prompt}.
\input{prompt/Client_prompt}

\subsection{Evaluator Simulator Prompt}
\label{subsec:evaluator}
We design an evaluator prompt to evaluate candidate responses from multiple perspectives and provide reliable preference signals. The evaluation considers three complementary roles: client, counselor, and judge, ensuring both empathetic quality and professional appropriateness. During automatic evaluation, we also adopt the same multi-perspective criteria in the LLM-as-a-Judge setting. The prompt design can be seen in Figure \ref{fig:evaluator_prompt}.
\input{prompt/evaluation_prompt}

\subsection{Counselor Simulator Prompt}
We design a stage-aware prompting framework that decomposes the counselor’s reasoning into three stages, each implemented with a dedicated prompt.
\paragraph{Counseling stage identification.}
The model infers the client’s current counseling phase based on the dialogue context. The prompt design can be seen in Figure \ref{fig:prompt1}.
\input{prompt/prompt1}

\paragraph{Psychological state assessment.}
The model evaluates the client’s psychological state in terms of distress, resilience, and motivation. The prompt design can be seen in Figure \ref{fig:prompt2}.
\input{prompt/prompt2}

\paragraph{Therapeutic strategy selection.}
The model selects appropriate counseling strategies conditioned on the inferred stage and psychological state. The prompt can be seen in Figure \ref{fig:prompt3_1} and Figure \ref{fig:prompt3_2}.
\input{prompt/prompt3_1}
\input{prompt/prompt3_2}

\section{Prompt Design for ESCRAG-R1 Training Framework}
\label{sec:appendix-prompt-for-training-and-inference}

% This section presents the prompt designs used in the ESCRAG-R1 training framework, including the reward model prompt and the policy model prompt.

\subsection{Reward Model Prompt}
\label{sec:appendix-reward-model-prompt}

The reward model is trained to evaluate whether a counselor response effectively addresses the client's emotional problem under a given conversation history.

\begin{tcolorbox}[breakable,title=Reward Model Prompt, colback=gray!5, colframe=gray!60]
\small
\textbf{Instruction:} You are a helpful assistant tasked with evaluating whether a patient's emotional problem has been effectively addressed following a conversation with a therapist.

\textbf{Conversation History:}

\{conversation\_history\}

\textbf{Candidate Response:}

\{candidate\_response\}
\end{tcolorbox}
\subsection{Policy Model Prompt}
\label{sec:appendix-policy-model-prompt}

The policy model is prompted to generate responses in a structured reasoning--response format. During retrieval-augmented generation, the model receives the dialogue history, an initial response, and retrieved dialogue exemplars from the external corpus, and then generates the final counselor response.

\begin{tcolorbox}[breakable,title=Policy Model Prompt, colback=gray!5, colframe=gray!60]
\small
\textbf{System Prompt:}

Now enter the role-playing mode. In the following conversation, you will play as a therapist in a counselling conversation with a patient. Your goal is to help the patient reduce their emotional distress and support them working through their challenges. You first think about the reasoning process in the mind and then provide the patient with the response. The reasoning process and response are enclosed within \texttt{<think> </think>} and \texttt{<response> </response>} tags, respectively, i.e., \texttt{<think> reasoning process here </think> <response> supportive response here </response>}.

\vspace{0.5em}
\textbf{User Prompt:}

You are an empathetic and helpful assistant. Given the following dialogue context, your initial response and response guidelines retrieved from an external library, your task is to generate a final response to the user. This response should incorporate empathy and understanding, provide helpful guidance or suggestions, and be conversational and natural.

\textbf{[Dialogue History]:}

\{dialogue\_history\}

\textbf{[Initial Response]:}

\{initial\_response\}

\textbf{[Retrieved Chunks from External Library]:}

\{retrieved\_dialogue\}
\end{tcolorbox}

\section{Parameter Setting}
\label{appendix-parameter-setting}

Table~\ref{tab:parameter-setting} summarizes the main parameter settings used for reward model training and Retrieval-Augmented GRPO. 
The reward model is built on Qwen-2.5-7B and trained using two A800 GPUs.
For policy optimization, \textsc{ESCRAG-R1}-3B is trained on two A800 GPUs, while \textsc{ESCRAG-R1}-7B is trained on four A800 GPUs.

\begin{table}[htbp]
\centering
\begin{tabular}{l l}
\hline
\textbf{Parameter} & \textbf{Value} \\
\hline
\multicolumn{2}{l}{\textit{Reward Model Training}} \\

Epochs & 5 \\
Per-device batch size & 1 \\
Gradient accumulation steps & 16 \\
Learning rate & $5\times10^{-5}$ \\
Reward centering coefficient & 0.01 \\
\hline
\multicolumn{2}{l}{\textit{Retrieval-Augmented GRPO}} \\

Batch size & 128 \\
Epochs & 2 \\
Learning rate & $1\times10^{-6}$ \\
KL coefficient & 0.05 \\
Number of sample generation & 4 \\
Number of retrieved chunks & 3 \\
Temperature & 0.9 \\
Top-$p$ & 0.95 \\
\hline
\end{tabular}
\caption{Hyperparameter settings for reward model training and Retrieval-Augmented GRPO.}
\label{tab:parameter-setting}
\end{table}

\section{Retrieval-Augmented GRPO and RAG Inference}
\label{subsec:ragrpo-and-rag-inference}

Algorithm~\ref{alg:ragrpo} summarizes how retrieval is used in ESCRAG-R1 during both training and inference. In retrieval-augmented GRPO, retrieved counseling exemplars are incorporated into the rollout process, and the reward of retrieval-guided responses is used to update the policy. In RAG inference, the optimized policy uses the same retrieval-augmented generation process to produce the final response, but no parameter update is performed. Thus, retrieval serves as a learning signal during training and as generation guidance during inference.

\input{table/algorithm}

\section{Human Evaluation}

To complement automatic evaluation, we conduct a human evaluation to assess the quality of generated responses. 
We recruit three students with backgrounds in psychology as human annotators and provide them with training before the annotation process. 
During training, annotators are introduced to the evaluation criteria and the scoring prompt used in our study, ensuring that they understand the meaning of each evaluation dimension and apply the criteria consistently.

For each dialogue context, annotators are presented with a blind pair of responses generated by different models, where the model identities are hidden to reduce potential bias. 
Following the evaluation prompt, annotators score each response from the Client--Counselor--Judge perspectives, covering emotional support quality, therapeutic coherence, and contextual appropriateness. 
After annotation, we compare the scores assigned to each response in the blind pair and determine model preference according to the higher overall score.

%% file: prompt/Client_prompt.tex
\begin{figure}[htbp]
\centering
\includegraphics[width=0.45\textwidth]{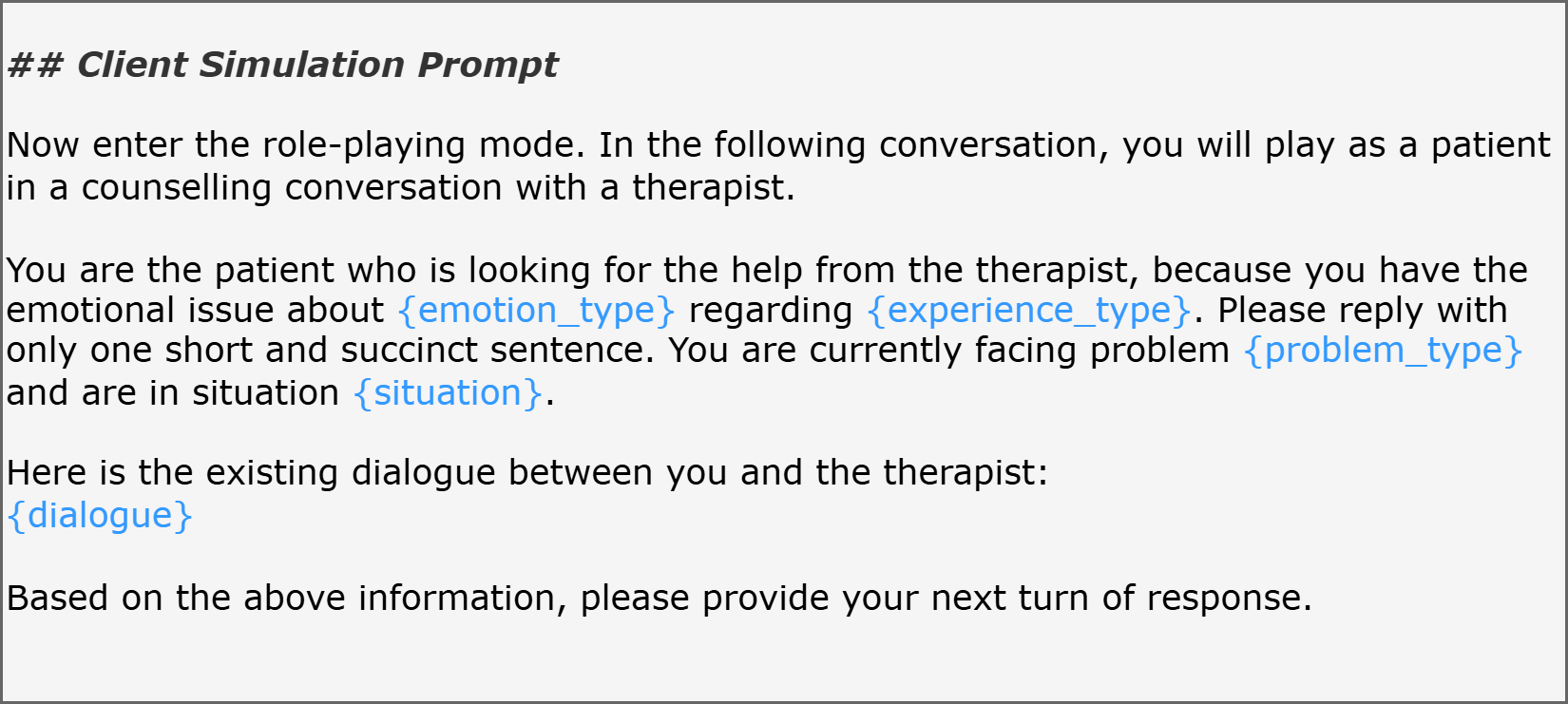}
\caption{Client role-playing prompt.}
\label{fig:client_prompt}
\end{figure}

%% file: prompt/evaluation_prompt.tex
\begin{figure}[htbp]
\centering
\includegraphics[width=0.45\textwidth]{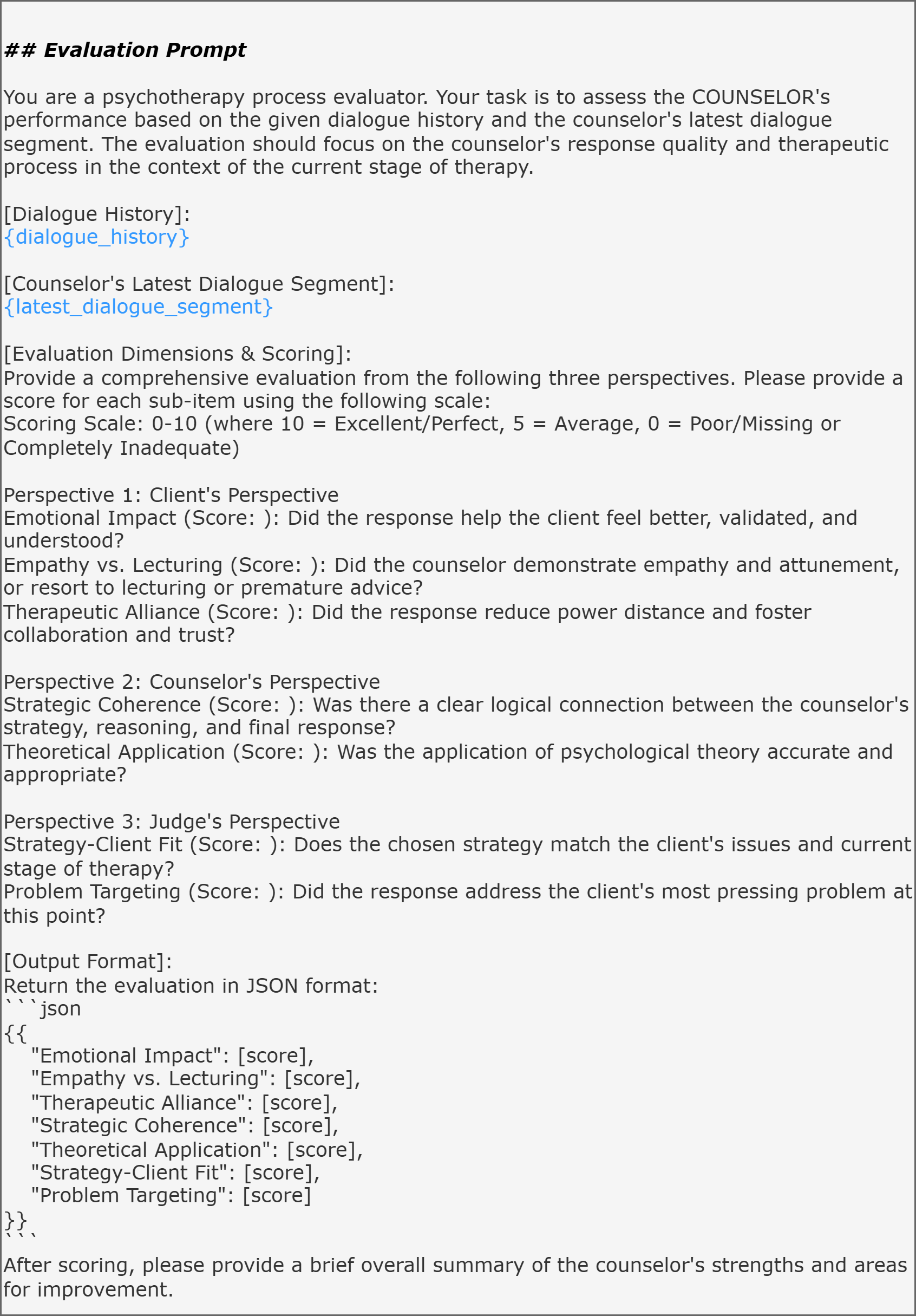}
\caption{Evaluator role-playing prompt.}
\label{fig:evaluator_prompt}
\end{figure}

%% file: prompt/prompt1.tex
\begin{figure}[htbp]
\centering
\includegraphics[width=0.45\textwidth]{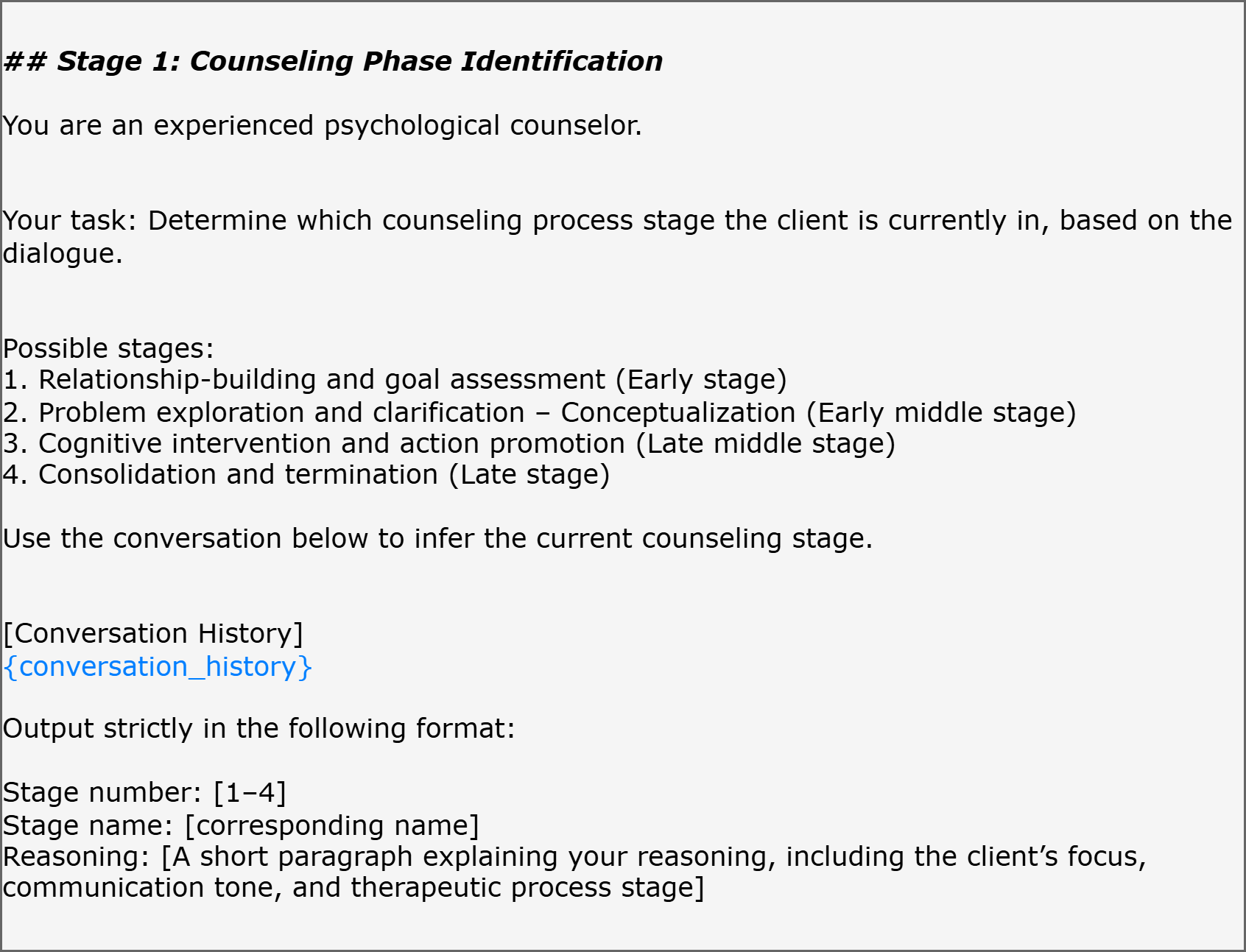}
\caption{Counselor role-playing prompt for counseling stage identification.}
\label{fig:prompt1}
\end{figure}

%% file: prompt/prompt2.tex
\begin{figure}[htbp]
\centering
\includegraphics[width=0.45\textwidth]{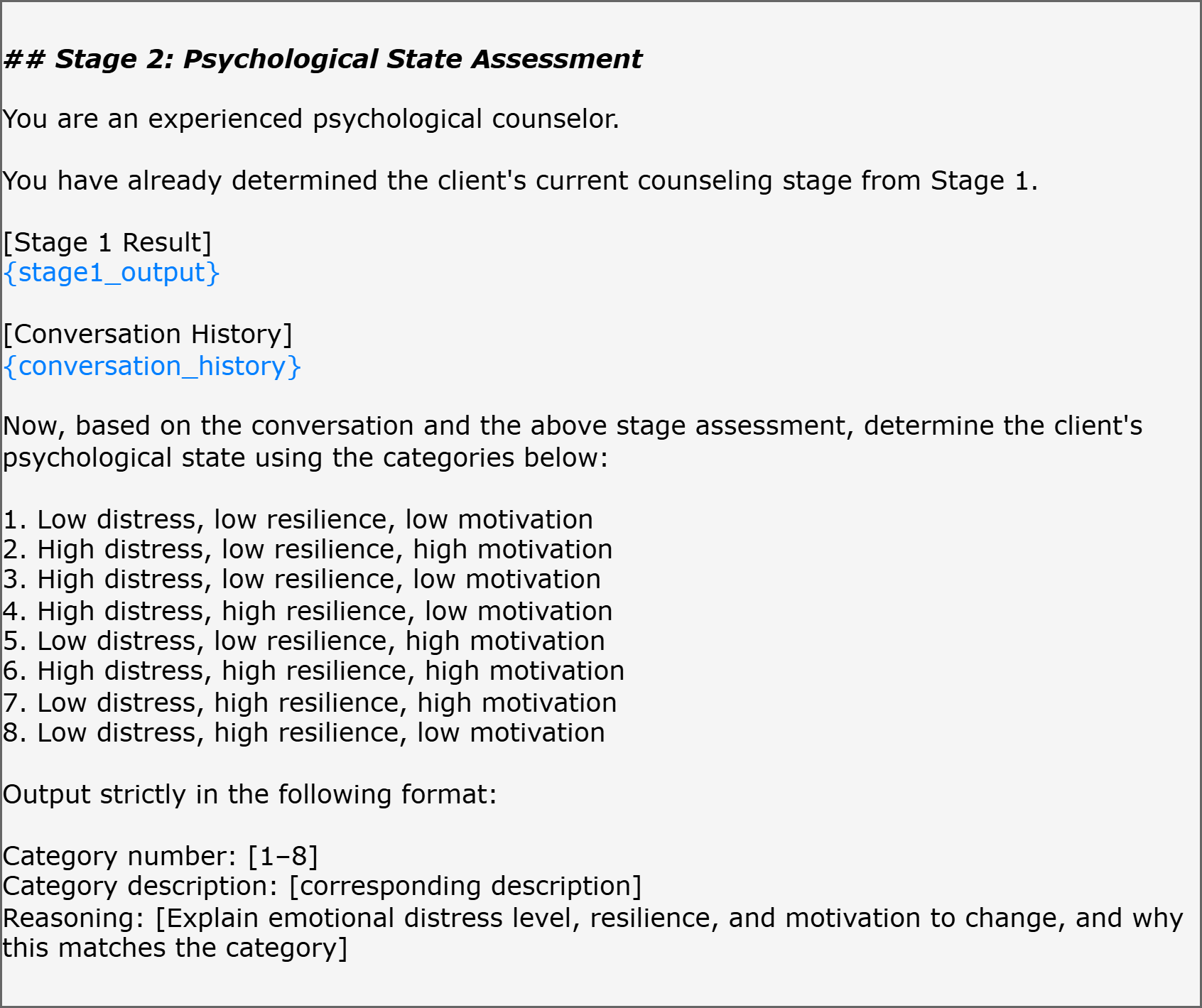}
\caption{Counselor role-playing prompt for psychological state assessment.}
\label{fig:prompt2}
\end{figure}

%% file: prompt/prompt3_1.tex
\begin{figure}[htbp]
\centering
\includegraphics[width=0.45\textwidth]{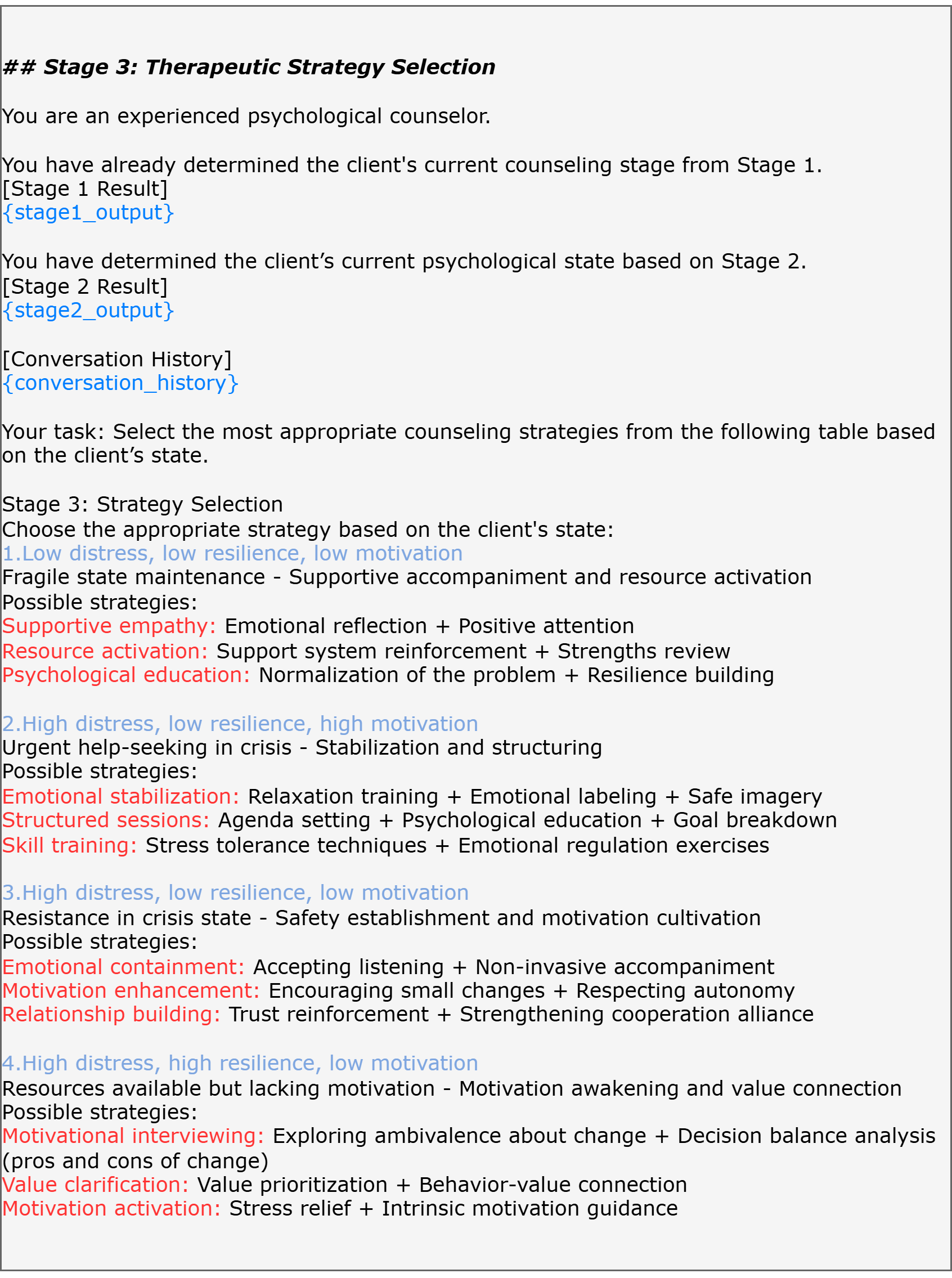}
\caption{Counselor role-playing prompt for therapeutic strategy selection (part 1).}
\label{fig:prompt3_1}
\end{figure}

%% file: prompt/prompt3_2.tex
\begin{figure}[htbp]
\centering
\includegraphics[width=0.45\textwidth]{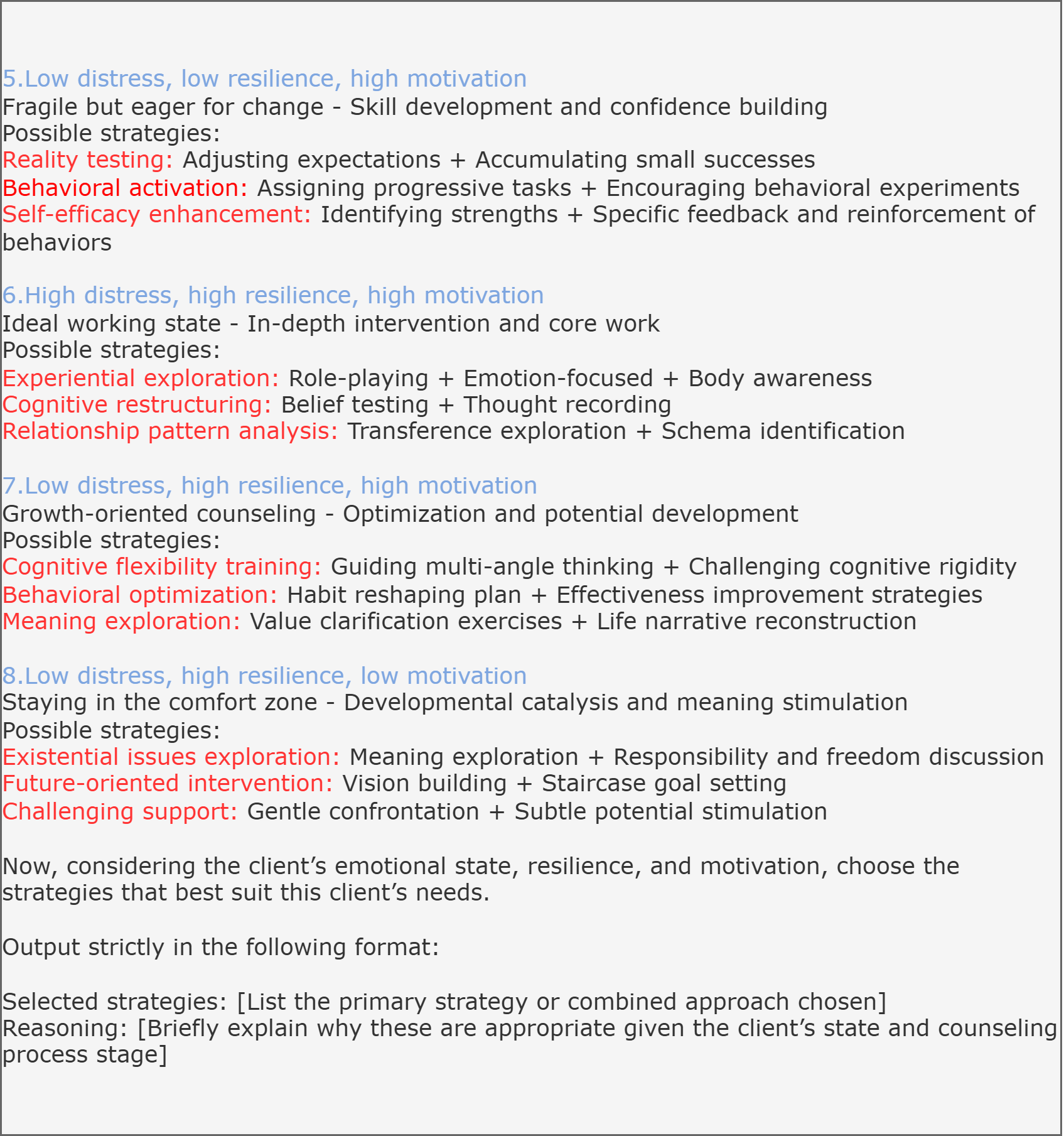}
\caption{Counselor role-playing prompt for therapeutic strategy selection (part 2).}
\label{fig:prompt3_2}
\end{figure}

%% file: table/algorithm.tex
% \begin{algorithm}[!t]
% \setlength\abovecaptionskip{0.2\baselineskip}
% \setlength\belowcaptionskip{0.2\baselineskip}
% \caption{Training Framework of ESCRAG-R1}
% \label{alg:ragrpo}
% \small
% \begin{algorithmic}[1]
% \Require Training set $\mathcal{D}=\{\mathcal{H}_i\}_{i=1}^{M}$, policy model $\pi_\theta$, reference policy $\pi_{\theta_{\mathrm{old}}}$, reward model $R_\phi$, retriever $\mathcal{R}$, group size $N$
% \Ensure Optimized policy parameters $\theta$

% \For{each training step}
%     \State Sample a dialogue state $\mathcal{H}_i$ from $\mathcal{D}$
%     \For{$j=1$ to $N$}
%         \State Generate an initial response $\hat{c}_i^j \sim \pi_\theta(\cdot \mid \mathcal{H}_i)$
%         \State Retrieve relevant exemplars $\mathcal{E}_i^j \leftarrow \mathcal{R}(\hat{c}_i^j)$
%         \State Generate a refined response $c_i^j \sim \pi_\theta(\cdot \mid \mathcal{H}_i,\hat{c}_i^j,\mathcal{E}_i^j)$
%         \State Compute reward $r_i^j \leftarrow R_\phi(\mathcal{H}_i, c_i^j)$
%     \EndFor
%     \State Compute group-relative advantages $\{A_i^j\}_{j=1}^{N}$ from rewards $\{r_i^j\}_{j=1}^{N}$
%     \State Compute policy ratios $\{\rho_i^j\}_{j=1}^{N}$ with respect to $\pi_{\theta_{\mathrm{old}}}$
%     \State Compute the clipped GRPO loss $\mathcal{L}_{\mathrm{GRPO}}(\theta)$
%     \State Update policy parameters $\theta$ by optimizing $\mathcal{L}_{\mathrm{GRPO}}(\theta)$
  
% \EndFor

% \State \Return $\theta$
% \end{algorithmic}
% \end{algorithm}
\begin{algorithm}[htbp]
\setlength\abovecaptionskip{0.2\baselineskip}
\setlength\belowcaptionskip{0.2\baselineskip}
\caption{Retrieval-Augmented GRPO and RAG Inference of ESCRAG-R1}
\label{alg:ragrpo}
\small
\begin{algorithmic}[1]
\Require Training set $\mathcal{D}_{\mathrm{train}}=\{\mathcal{H}_i\}_{i=1}^{M}$, test set $\mathcal{D}_{\mathrm{test}}=\{\mathcal{H}_i\}_{i=1}^{M'}$, policy model $\pi_\theta$, old policy $\pi_{\theta_{\mathrm{old}}}$, reward model $R_\phi$, retriever $\mathcal{R}$, group size $N$
\Ensure Optimized policy parameters $\theta$ and final responses $\{c_i^{\mathrm{final}}\}_{i=1}^{M'}$

\Statex \textbf{Retrieval-Augmented GRPO}
\For{each training step}
    \State Sample a dialogue state $\mathcal{H}_i$ from $\mathcal{D}_{\mathrm{train}}$
    \For{$j=1$ to $N$}
        \State Generate an initial response $\hat{c}_i^j \sim \pi_\theta(\cdot \mid \mathcal{H}_i)$
        \State Retrieve relevant exemplars $\mathcal{E}_i^j \leftarrow \mathcal{R}(\hat{c}_i^j)$
        \State Generate a retrieval-augmented response
        $c_i^j \sim \pi_\theta(\cdot \mid \mathcal{H}_i,\hat{c}_i^j,\mathcal{E}_i^j)$
        \State Compute reward $r_i^j \leftarrow R_\phi(\mathcal{H}_i, c_i^j)$
    \EndFor
    \State Compute group-relative advantages $\{A_i^j\}_{j=1}^{N}$ from rewards $\{r_i^j\}_{j=1}^{N}$
    \State Compute importance sampling ratios $\{\rho_i^j\}_{j=1}^{N}$ with respect to $\pi_{\theta_{\mathrm{old}}}$
    \State Compute the clipped GRPO loss $\mathcal{L}_{\mathrm{GRPO}}(\theta)$
    \State Update policy parameters $\theta$ by optimizing $\mathcal{L}_{\mathrm{GRPO}}(\theta)$
\EndFor
\State Obtain the optimized policy $\pi_\theta$

\Statex \textbf{RAG Inference}
\For{each dialogue state $\mathcal{H}_i$ in $\mathcal{D}_{\mathrm{test}}$}
    \State Generate an initial response $\hat{c}_i \sim \pi_\theta(\cdot \mid \mathcal{H}_i)$
    \State Retrieve relevant exemplars $\mathcal{E}_i \leftarrow \mathcal{R}(\hat{c}_i)$
    \State Generate the final response
    $c_i^{\mathrm{final}} \sim \pi_\theta(\cdot \mid \mathcal{H}_i,\hat{c}_i,\mathcal{E}_i)$
\EndFor

\State \Return $\theta$ and $\{c_i^{\mathrm{final}}\}_{i=1}^{M'}$
\end{algorithmic}
\end{algorithm}